\documentclass{article}

    \PassOptionsToPackage{numbers, compress}{natbib}
 \usepackage[preprint]{neurips_2026}

\usepackage[utf8]{inputenc} 
\usepackage[T1]{fontenc}    
\usepackage{hyperref}       
\usepackage{url}            
\usepackage{booktabs}       
\usepackage{amsfonts}       
\usepackage{nicefrac}       
\usepackage{microtype}      
\usepackage{xcolor}         

\usepackage{algorithm}
\usepackage{algorithmic}
\usepackage{amsmath} 
\usepackage{varwidth}
\usepackage{graphicx}
\usepackage{tabularx}
\usepackage{multirow}
\usepackage{array}
\usepackage{subcaption}

\title{Decision-Aware Attention Propagation for Vision Transformer Explainability}

\author{%
  Sehyeong Jo\thanks{Authors contributed equally to this paper.} \\  
  Aarhus University\\  
  \texttt{sehyeong.jo@agro.au.dk} \\
  \And
  Gangjae Jang\footnotemark[1] \\  
  University of Colorado Boulder\\  
  \texttt{gangjae.jang@colorado.edu} \\
  \And
  Haesol Park\\
  KIST\\
  \texttt{haesol@kist.re.kr} \\
}

\begin{document}

\maketitle

\begin{abstract}
Vision Transformers (ViTs) have become a dominant architecture in computer vision, yet their prediction process remains difficult to interpret because information is propagated through complex interactions across layers and attention heads. Existing attention based explanation methods provide an intuitive way to trace information flow. However, they rely mainly on raw attention weights, which do not explicitly reflect the final decision and often lead to explanations with limited class discriminability. In contrast, gradient based localization methods are more effective at highlighting class specific evidence, but they do not fully exploit the hierarchical attention propagation mechanism of transformers. To address this limitation, we propose Decision-Aware Attention Propagation (DAP), an attribution method that injects decision-relevant priors into transformer attention propagation. By estimating token importance through gradient based localization and integrating it into layer wise attention rollout, the method captures both the structural flow of attention and the evidence most relevant to the final prediction. Consequently, DAP produces attribution maps that are more class sensitive, compact, and faithful than those generated by conventional attention based methods. Extensive experiments across Vision Transformer variants of different model scales show that DAP consistently outperforms existing baselines in both quantitative metrics and qualitative visualizations, indicating that decision aware propagation is an effective direction for improving ViT interpretability.
\end{abstract}

\section{Introduction}
\label{sec:intro}

Vision Transformer (ViT) has become one of the most influential architectures in modern computer vision~\cite{dosovitskiy2020image}. By modeling images as sequences of tokens and leveraging self-attention to capture long-range dependencies, ViT-based models have achieved strong performance across a wide range of vision tasks. As these models are increasingly adopted in practical applications, understanding how they make predictions has become an important problem~\cite{han2022survey, huang2022visual}. Beyond predictive accuracy, interpretability is essential for improving model transparency, diagnosing failure cases, and building trust in model decisions.

Despite the increasing demand for interpretability, many existing explainable AI (XAI) methods were originally developed for convolutional neural networks (CNNs)~\cite{ibrahim2023explainable, nazir2024utilizing}. Such methods have been highly effective for CNN-based architectures, where local receptive fields and hierarchical convolutional features provide a natural basis for explanation. However, directly transferring these approaches to transformer-based vision models is not always sufficient. Unlike CNNs, ViTs rely on token-wise interactions and layer-wise attention propagation, which play a central role in how information is aggregated and how final predictions are formed~\cite{khan2023survey}. Therefore, an explanation method for ViTs should account for these transformer-specific characteristics rather than simply reuse explanation strategies designed for convolutional architectures.

Existing approaches for explaining ViTs can be broadly grouped into attention-based and gradient-based methods~\cite{kashefi2026explainability}. Attention-based methods are naturally aligned with transformer architectures because they preserve token interactions and layer-wise information flow. However, attention alone is not optimized to separate evidence for a specific target class, which often leads to explanations that remain diffuse or weakly class-discriminative~\cite{palikhe2025towards}. By contrast, gradient-based methods more directly capture class-relevant evidence, but they do not explicitly model how relevance propagates through the attention structure of the transformer~\cite{arrieta2020explainable}. As a result, existing approaches often prioritize either propagation faithfulness or class discrimination, rather than resolving both simultaneously~\cite{stassin2023explainability}. Recent studies have attempted to combine attention propagation with gradient-derived decision cues, suggesting that transformer-aware structure and class-specific localization can be complementary~\cite{KAZMIERCZAK2025103184}. However, many such approaches incorporate gradient information only as head-level reweighting or auxiliary guidance, rather than directly controlling how token-to-token relevance is propagated across layers. Consequently, they may improve class awareness only partially, while leaving the underlying propagation process weakly discriminative or overly diffuse.

\begin{figure}
  \centering
  \includegraphics[width=\textwidth]{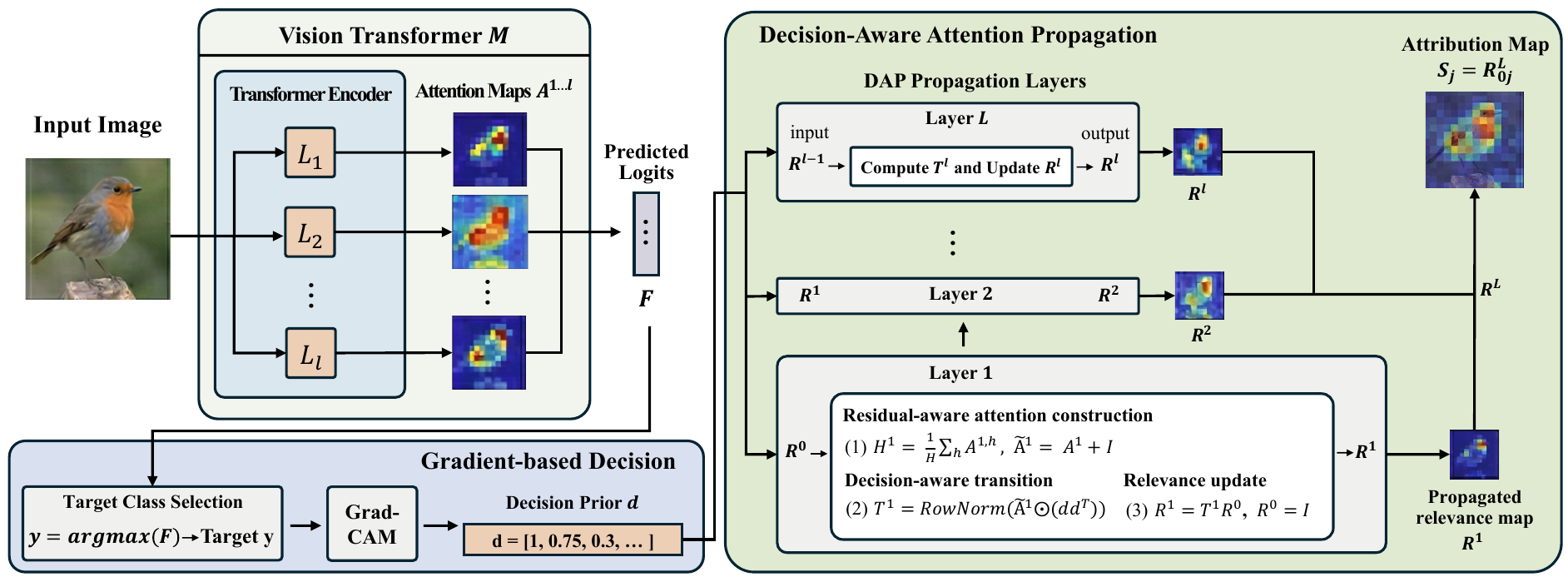}  
  \caption{Overall pipeline of Decision-Aware Attention Propagation (DAP)}
  \label{fig:overview}
\end{figure}

To address this limitation, we propose Decision-Aware Attention Propagation (DAP), which injects gradient-derived decision cues into the propagation operator of transformer attention itself (Figure~\ref{fig:overview}). DAP first estimates token-level importance from a gradient-based localization signal and converts it into a decision prior. This prior is then used to modulate the token-to-token transition matrix at each layer, so that relevance is preferentially propagated through tokens that are more strongly associated with the target prediction. In contrast to methods that use gradients only for head weighting or post-hoc localization, DAP directly reshapes layer-wise relevance flow while preserving the residual-aware attention structure of the transformer. As a result, it produces explanations that are more class-discriminative, propagation-consistent, and better aligned with the model’s final decision.

\section{Related Work}
\label{sec:rel}

\paragraph{Attention Based Attribution Methods}
Attention based attribution methods explain Vision Transformer predictions by tracing how attention propagates across layers. Attention Rollout~\cite{abnar2020quantifying} aggregates attention matrices through the network to estimate token level relevance, providing an intuitive view of information flow. Attribution Rollout~\cite{xu2023attribution} extends this idea by incorporating attribution related signals into the rollout process, making the resulting explanations more aligned with token importance than raw attention alone. Chefer et al.~\cite{chefer2021transformer} further improve transformer interpretability by introducing a relevance propagation framework that produces class specific explanations beyond simple attention visualization. More recently, GMAR~\cite{jo2025gmar} weights individual attention heads using gradient-based importance scores before rollout, allowing more informative heads to contribute more strongly to the final attribution map. While these methods effectively preserve transformer attention structure, gradient cues in existing hybrid variants are often used only for head reweighting or auxiliary scoring, rather than for directly shaping token-level relevance propagation. Consequently, the resulting explanations may remain diffuse or only partially class-discriminative.

\paragraph{Gradient Based Localization Methods}
Gradient based localization methods explain model predictions by identifying input regions that most strongly influence the target output. CAM~\cite{zhou2016learning} first localized class relevant regions using weighted feature maps from convolutional networks, and Grad-CAM~\cite{selvaraju2017grad} generalized this idea through class gradients to highlight prediction relevant spatial regions. Grad-CAM++~\cite{chattopadhay2018grad} further improved this framework by providing better localization for multiple instances and fine grained visual evidence. CDAM~\cite{brocki2023class} extends this direction by introducing class discriminative attention maps that better emphasize prediction specific evidence. AGCAM~\cite{leem2024attention} combines gradient information with self attention to generate explanations that are both semantically meaningful and spatially localized for Vision Transformers. LeGrad~\cite{bousselham2025legrad} further adapts gradient based explanation to ViTs by computing gradients with respect to attention maps and aggregating them across layers to identify influential image regions. Although these methods capture class-specific evidence effectively, they are not explicitly constrained to follow transformer attention propagation, making it difficult to reflect how relevance evolves through layer-wise token interactions.

Taken together, these observations suggest that the remaining challenge is not simply to combine gradients with attention, but to determine how decision cues should shape relevance propagation itself. In particular, directly injecting decision cues into the token-level propagation operator remains less explored than head-level reweighting or post-hoc localization.

\section{Method}
\label{sec:method}

DAP is proposed to improve the interpretability of ViTs by integrating class-specific decision cues into the attention propagation process. Unlike conventional attention-based methods, which preserve transformer-specific information flow but may lack discriminative focus, DAP introduces gradient-derived token importance as a decision prior to guide propagation toward prediction-relevant evidence. Specifically, the method first estimates the contribution of each token to the target class using a gradient-based localization signal, and then incorporates this prior into the layer-wise accumulation of attention. As a result, DAP preserves the propagation characteristics of transformer attention while encouraging decision-relevant tokens to contribute more strongly to the final attribution map. Rather than using gradients only to score attention heads or produce a separate localization map, DAP modulates the token-to-token propagation matrix itself with decision priors. This allows class-specific evidence to influence how relevance flows across layers, while preserving the transformer’s propagation structure.

\subsection{Preliminaries}

Let \(x \in \mathbb{R}^{H \times W \times C}\) denote an input image. In a ViT, \(x\) is represented as a sequence of \(P\) patch tokens together with a class token, resulting in \(P+1\) tokens in total. For each layer \(l \in \{1,\dots,L\}\) and head \(h \in \{1,\dots,H\}\), let \(A^{l,h} \in \mathbb{R}^{(P+1)\times(P+1)}\) denote the self-attention matrix. The attention maps are aggregated across heads as
\begin{equation}
A^l = \frac{1}{H}\sum_{h=1}^{H} A^{l,h}.
\end{equation}
To account for residual connections, we define a residual-aware transition matrix as
\begin{equation}
\tilde{A}^l = A^l + I,
\end{equation}
where \(I\) is the identity matrix. Based on this formulation, token relevance can be propagated across layers by recursively composing layer-wise transition matrices. Let \(R \in \mathbb{R}^{(P+1)\times(P+1)}\) denote the accumulated relevance matrix, initialized as \(R = I\). After propagation through the transformer layers, each element \(R_{ij}\) represents the accumulated contribution from token \(j\) on token \(i\). Under this convention, the class-token row \(R_{0,:}\) measures how strongly each patch token contributes to the class token after layer-wise propagation. Therefore, the final attribution score for patch token \(j\) is obtained from \(R_{0j}\).

\subsection{Decision-Aware Attention Propagation}

DAP incorporates gradient-derived decision cues into layer-wise attention propagation in order to guide relevance accumulation toward tokens that are more strongly associated with the target prediction. Let \(d \in \mathbb{R}^{P+1}\) denote a decision prior over tokens. Given the residual-aware transition matrix $\tilde{A}^l$ at layer $l$, DAP defines a decision-aware transition matrix by modulating each token-to-token interaction with the corresponding prior values:
\begin{equation}
\hat{T}_{ij}^l = \tilde{A}_{ij}^l d_i d_j .
\end{equation}
The multiplicative modulation $d_i d_j$ makes each token-to-token transition decision-aware by emphasizing interactions between source and target tokens that are both relevant to the target prediction. This pairwise form favors transitions only when both sides of the interaction carry strong decision evidence, rather than biasing propagation with a purely local score on only one token. In this way, the decision prior shapes the propagation process itself instead of being applied only as a post-hoc weighting on the final attribution map. When the prior is uniform, the formulation can be viewed as reducing to the residual-aware propagation used in standard attention rollout. In this work, the decision prior is computed once for the selected target class and shared across all transformer layers, so that the comparison isolates how decision cues affect propagation rather than introducing layer-specific prior estimation noise.
The resulting matrix is then row-normalized to preserve a valid layer-wise propagation structure:
\begin{equation}
T_{ij}^l = \frac{\hat{T}_{ij}^l}{\sum_k \hat{T}_{ik}^l}.
\end{equation}
This formulation allows DAP to preserve the transformer attention structure while introducing class-discriminative guidance into relevance propagation.

\subsubsection{Gradient-based Decision Prior}

To estimate decision-relevant token importance, DAP first computes a gradient-based localization signal for the target prediction. Given the model output \(F\) for input image \(x\), the target class is determined as
\begin{equation}
y = \arg \max\limits_{c} F_c(x).
\end{equation}
A token importance vector \(s = [s_1, \dots, s_P]\) is obtained using Grad-CAM and normalized by min-max scaling to define the decision prior:
\begin{equation}
d = \left[1,\; \frac{s_1-\min(s)}{\max(s)-\min(s)},\; \dots,\; \frac{s_P-\min(s)}{\max(s)-\min(s)}\right].
\end{equation}
Here, the class token is assigned a fixed value of \(1\), while patch tokens are weighted according to their gradient-derived relevance to the predicted class. The class-token prior is fixed to 1 so that propagation remains anchored to the prediction token while only patch-token contributions are modulated by the decision cue. The resulting patch-level scores are aligned with the token grid and used as the decision prior over patch tokens.

\subsubsection{Attention Flow Preserving Propagation}

Using the decision-aware transition matrix \(T^l\), relevance is recursively propagated across layers as
\begin{equation}
R^{(l)} = T^l R^{(l-1)}, \qquad R^{(0)} = I.
\end{equation}
After the final layer, the attribution score for each patch token is obtained from the class-token row of the propagated relevance matrix:
\begin{equation}
S_j = R^{(L)}_{0j}, \qquad j = 1,\dots,P.
\end{equation}
In this way, DAP maintains the layer-wise attention flow of the transformer while emphasizing transitions between tokens that are more relevant to the target prediction. Because this modulation is applied to the residual-augmented attention matrix and followed by row normalization, DAP preserves the layer-wise transition semantics of transformer attention while steering relevance flow toward decision-relevant evidence.

\begin{center}
\begin{varwidth}{0.8\columnwidth}

\begin{algorithm}[H]
\caption{Decision-Aware Attention Propagation (DAP)}
\label{alg:dap}

\begin{algorithmic}[1]

\REQUIRE Vision Transformer $M$, input image $x$
\ENSURE Attribution heatmap $H$
\STATE $M.\mathrm{zero\_grad}()$
\STATE $F \leftarrow M(x)$
\STATE $y \leftarrow \arg\max_c F_c$
\STATE Compute patch-token importance $s$ using Grad-CAM for class $y$
\STATE $d_i \leftarrow \frac{s_i-\min(s)}{\max(s)-\min(s)}$
\STATE $d \leftarrow [1,d_1,\dots,d_P]$
\STATE Extract attention layers $\{A^{1,h},\dots,A^{L,h}\}$
\STATE $R \leftarrow I$
\FOR{$l=1$ to $L$}
    \STATE $A^l \leftarrow \frac{1}{H}\sum_h A^{l,h}$
    \STATE $\tilde{A}^l \leftarrow A^l + I_{P+1}$
    \STATE $\hat{T}^l \leftarrow \tilde{A}^l \odot (d d^\top)$
    \STATE $T^l \leftarrow \mathrm{RowNorm}(\hat{T}^l)$
    \STATE $R^{(l)} \leftarrow T^l R^{(l-1)}$
\ENDFOR
\STATE $S \leftarrow R^{(L)}_{0,1:P}$
\STATE $H \leftarrow \mathrm{Interp}(\mathrm{ReshapeToGrid}(S))$
\RETURN $H$

\end{algorithmic}
\end{algorithm}

\end{varwidth}
\end{center}

\subsection{Detailed Procedure of Decision-Aware Attention Propagation}

Algorithm~\ref{alg:dap} summarizes the overall DAP procedure. Given an input image, DAP first estimates a gradient-based decision prior over tokens for the selected target class. This prior is injected into the residual-aware attention transitions at every layer, so that relevance is propagated through token interactions that are more strongly associated with the final prediction. The resulting decision-aware transitions are recursively composed across layers, and the class-token row of the final relevance matrix is reshaped and interpolated to obtain the attribution heatmap.

\section{Experiment}
\label{sec:exp}

\subsection{Experimental Setup}

Experiments were conducted on randomly sampled images from the ImageNet-1K~\cite{deng2009imagenet} training split using pretrained ViT backbones~\cite{steiner2021augreg, rw2019timm}. To evaluate the proposed method across model scales, we considered multiple ViT variants, and generated each explanation map with respect to the predicted class. All methods were evaluated under a unified PyTorch pipeline with identical backbone-specific preprocessing and heatmap conversion for fair comparison. Results are reported under two sampling settings: a non-balanced setting with 800 randomly selected images, and a class-balanced setting with 800 images sampled from 200 classes with 4 images per class. For CS evaluation, the alternative class was chosen as the highest-scoring non-predicted class for each input, so that class sensitivity was measured against a strong competing decision. For top-$k$ based evaluation, $k$ was fixed to 10\% of the patch tokens. In TCC, only the top-10\% patches were retained and the remaining regions were zero-masked after patch-to-image upsampling, while the same ratio was used in AFS to compute top-$k$ mass. All experiments were conducted on a single NVIDIA GeForce RTX 5060 Ti GPU.

\subsection{Evaluation Metrics}
We evaluate explanation quality using both perturbation-based and attribution-based metrics. Deletion and Insertion assess how the model confidence changes when salient regions are progressively removed from or inserted into the input. In general, better explanations yield lower Deletion and higher Insertion scores. We further consider four attribution-based metrics that capture class discrimination, decision consistency, compactness, and layer-wise alignment in token-level explanations.

\paragraph{Class Sensitivity (CS).}
CS measures how differently an explanation responds to the predicted class and to an alternative class. Let $E_{\text{pred}}$ and $E_{\text{alt}}$ denote the corresponding explanation maps. We define
\begin{equation}
\mathrm{CS} = 1 - \frac{\rho_s(E_{\text{pred}}, E_{\text{alt}})}{2},
\end{equation}
where $\rho_s$ is the Spearman rank correlation. Higher CS indicates stronger class discrimination.

\paragraph{Token Contribution Consistency (TCC).}
TCC evaluates whether the top-$k$ patches selected by an explanation are sufficient to preserve the target prediction. After retaining only the selected patches, we compare the target-class confidence before and after masking:
\begin{equation}
\mathrm{TCC} = \frac{p(y_t \mid x_{\text{top}k})}{p(y_t \mid x)}.
\end{equation}
Higher TCC indicates that the selected tokens preserve more of the evidence used for the original decision.

\paragraph{Attention Flow Sparsity (AFS).}
AFS measures how strongly the explanation mass is concentrated on a small subset of patches. In this work, we quantify it using top-$k$ mass:
\begin{equation}
\mathrm{AFS}_{\text{top}k} =
\frac{\sum_{i \in \mathrm{TopK}(E)} E_i}{\sum_j E_j}.
\end{equation}
Higher AFS indicates a more compact explanation.

\paragraph{Layer-wise Decision Alignment (LDA).}
LDA evaluates how well an intermediate layer-wise relevance map aligns with the final attribution map produced by the same method. For the layer-$l$ relevance map $M_l$ and reference map $M_{\mathrm{ref}}$, we compute
\begin{equation}
\mathrm{LDA}_l = \rho_s(M_l, M_{\mathrm{ref}}),
\end{equation}
where $\rho_s$ is the Spearman rank correlation. Higher LDA indicates stronger consistency of layer-wise propagation with the final decision map.

\section{Result}
\label{sec:result}

\subsection{Quantitative Evaluation}

In the quantitative evaluation, we perform two complementary comparisons to assess the effectiveness of the proposed method. The approach is evaluated against representative attention-based methods to establish a direct performance comparison within attention-driven explanation frameworks. In addition, to examine the role of gradient information, we compare it with GMAR and include Grad-CAM as a control baseline. This design enables a comprehensive evaluation of the method both as an attention-based explanation approach and as a framework that integrates gradient-derived decision cues into attention propagation.

\subsubsection{Comparison with Attention-based Methods}

The comparison with attention-based methods (Table ~\ref{tab:metric_ar}) shows that the proposed approach behaves differently across ViT scales. Its effect was limited on ViT-T, suggesting that DAP is less effective when the backbone has weaker representations and less reliable attention structure. In contrast, the gains became clearer as model size increased. On ViT-S, the proposed method achieved the best performance on Ins, CS, TCC, and LDA in both settings. On ViT-B and ViT-L, it also showed favorable results across most metrics, reaching 0.755/0.259/0.633 on ViT-B and 0.798/0.355/0.733 on ViT-L under the balanced setting in terms of Ins, CS, and TCC, respectively. The CS scores of AR and GMAR were extremely small across all settings and therefore appeared as 0.000 after rounding. Overall, these results suggest that decision-aware propagation becomes more effective with larger backbones, which provide richer representations and more structured attention patterns.

Across the balanced and non-balanced settings, the compared methods showed only minor differences, with largely unchanged relative rankings and small score gaps overall. For DAP, TCC changed from 0.380 to 0.372 on ViT-S, from 0.633 to 0.673 on ViT-B, and from 0.733 to 0.756 on ViT-L. These results suggest that class balance has limited effect on the comparative behavior of attention-based explanation methods, whereas backbone scale has a more noticeable influence, likely because explanation quality depends more on the learned representations and attention structure than on the class composition of the evaluation set.

\renewcommand{\tabularxcolumn}[1]{m{#1}}
\begin{table}[!t]
    \centering
    \footnotesize
    \renewcommand{\arraystretch}{1.2}
    
    \begin{tabularx}{\textwidth}{c|c|*{6}{>{\centering\arraybackslash}X}|*{6}{>{\centering\arraybackslash}X}}
    \hline
    
    \multirow{2}{*}{\textbf{Model}} & 
    \multirow{2}{*}{\textbf{Method}} & 
    \multicolumn{6}{c|}{\textbf{Balanced Class}} & 
    \multicolumn{6}{c}{\textbf{Non-Balanced Class}} \\
    
    \cline{3-14}
    
    & & 
    \textbf{Del\,{\footnotesize$\downarrow$}} & 
    \textbf{Ins} & 
    \textbf{CS} & 
    \textbf{TCC} & 
    \textbf{AFS} & 
    \textbf{LDA} & 
    \textbf{Del\,{\footnotesize$\downarrow$}} &
    \textbf{Ins} & 
    \textbf{CS} & 
    \textbf{TCC} & 
    \textbf{AFS} & 
    \textbf{LDA} \\
    
    \hline
    
    \multirow{5}{*}{ViT-T}
    & AR    & 0.187 & 0.533 & 0.000 & 0.127 & 0.246 & 0.063 & 0.185 & 0.533 & 0.000 & \underline{0.133} & 0.242 & 0.058 \\
    & AttR  & \textbf{0.112} & \textbf{0.619} & 0.096 & \textbf{0.238} & 0.317 & 0.048 & \textbf{0.110} & \textbf{0.627} & \underline{0.102} & \textbf{0.222} & 0.318 & 0.048 \\
    & Chefer & \underline{0.162} & \underline{0.542} & \underline{0.097} & \underline{0.134} & \textbf{0.568} & - &\underline{0.161} & \underline{0.546} & 0.091 & 0.130 & \textbf{0.566} & - \\
    & GMAR  & 0.185 & 0.531 & 0.000 & 0.123 & 0.245 & \underline{0.067} & 0.182 & 0.531 & 0.000 & 0.125 & 0.240 & \underline{0.064} \\
    & \textbf{Ours} & 0.200 & 0.510 & \textbf{0.386} & 0.117 & \underline{0.525} & \textbf{0.273} & 0.200 & 0.519 & \textbf{0.392} & 0.107 & \underline{0.523} & \textbf{0.278} \\    
    \hline
    \multirow{5}{*}{ViT-S}
    & AR & 0.327 & 0.583 & 0.000 & 0.127 & 0.241 & 0.028 & 0.328 & 0.591 & 0.000 & 0.139 & 0.242 & 0.037 \\
    & AttR & \textbf{0.205} & \underline{0.692} & \underline{0.059} & \underline{0.326} & 0.340 & \underline{0.030} & \textbf{0.204} & \textbf{0.701} & \underline{0.064} & \underline{0.321} & 0.340 & 0.028 \\
    & Chefer & 0.243 & 0.643 & 0.058 & 0.253 & \textbf{0.453} & - & 0.246 & 0.654 & 0.059 & 0.272 & \underline{0.440} & - \\
    & GMAR & 0.329 & 0.578 & 0.000 & 0.121 & 0.244 & 0.027 & 0.330 & 0.586 & 0.000 & 0.137 & 0.245 & \underline{0.037} \\
    & \textbf{Ours} & \underline{0.209} & \textbf{0.696} & \textbf{0.306} & \textbf{0.380} & \underline{0.445} & \textbf{0.226} & \underline{0.206} & \underline{0.701} & \textbf{0.309} & \textbf{0.372} & \textbf{0.441} & \textbf{0.216} \\
    \hline
    \multirow{5}{*}{ViT-B}
    & AR & 0.397 & 0.654 & 0.000 & 0.273 & 0.229 & \underline{0.032} & 0.412 & 0.658 & 0.000 & 0.287 & 0.231 & 0.038 \\
    & AttR & \underline{0.269} & \underline{0.738} & \underline{0.046} & \underline{0.533} & 0.329 & 0.005 & \underline{0.273} & \underline{0.750} & \underline{0.044} & \underline{0.568} & 0.330 & 0.005 \\
    & Chefer & 0.309 & 0.703 & 0.027 & 0.437 & \textbf{0.451} & - & 0.310 & 0.721 & 0.024 & 0.476 & \textbf{0.444} & - \\
    & GMAR & 0.396 & 0.653 & 0.000 & 0.273 & 0.231 & 0.032 & 0.409 & 0.658 & 0.000 & 0.284 & 0.233 & \underline{0.040} \\
    & \textbf{Ours} & \textbf{0.249} & \textbf{0.755} & \textbf{0.259} & \textbf{0.633} & \underline{0.358} & \textbf{0.103} & \textbf{0.259} & \textbf{0.770} & \textbf{0.252} & \textbf{0.673} & \underline{0.353} & \textbf{0.116} \\
    \hline
    \multirow{5}{*}{ViT-L}
    & AR & 0.564 & 0.694 & 0.000 & 0.322 & 0.224 & \underline{0.053} & 0.570 & 0.699 & 0.000 & 0.322 & 0.225 & \underline{0.055} \\
    & AttR & \underline{0.436} & \underline{0.751} & 0.056 & \underline{0.501} & \underline{0.295} & 0.023 & \underline{0.441} & \underline{0.751} & 0.052 & \underline{0.507} & \underline{0.293} & 0.020 \\
    & Chefer & 0.609 & 0.632 & \underline{0.092} & 0.210 & 0.238 & - & 0.613 & 0.636 & \underline{0.094} & 0.234 & 0.245 & - \\
    & GMAR & 0.565 & 0.686 & 0.000 & 0.292 & 0.225 & 0.049 & 0.568 & 0.692 & 0.000 & 0.299 & 0.226 & 0.051 \\
    & \textbf{Ours} & \textbf{0.398} & \textbf{0.798} & \textbf{0.355} & \textbf{0.733} & \textbf{0.344} & \textbf{0.100} & \textbf{0.398} & \textbf{0.795} & \textbf{0.367} & \textbf{0.756} & \textbf{0.342} & \textbf{0.100} \\
    \hline
    
    \end{tabularx}
    \vspace{3pt}
    \caption{Comparison of the proposed method with attention-based XAI approaches across multiple ViT models under both balanced and non-balanced class settings}
    \label{tab:metric_ar}
\end{table}

\subsubsection{Ablation on Decision Prior Injection}

We ablate how the decision prior is incorporated into propagation on the balanced setting using ViT-B and ViT-L. We compare Uniform, which uses the standard transition without prior modulation, Target-only, which applies the prior only to the receiving token, Final-only, which applies the prior only after propagation, and the proposed Pairwise modulation. We also tested Source-only, but it produced identical results to Pairwise. Under row normalization, the target-side factor in the pairwise form, $\hat{T}_{ij}^{\,l}=\tilde{A}_{ij}^{\,l} d_i d_j$, is canceled within each row, making it algebraically equivalent to Source-only, $\hat{T}_{ij}^{\,l}=\tilde{A}_{ij}^{\,l} d_j$, in the current propagation scheme.

\begin{table}[ht]
    \centering
    \renewcommand{\arraystretch}{1.2}
    
    \begin{tabular}{c|c|c c c c}
    \hline
    
    \textbf{Model} & 
    \textbf{Variant} & 
    \textbf{CS} & 
    \textbf{TCC} & 
    \textbf{LDA} & 
    \textbf{AFS} \\
    
    \hline
    
    \multirow{4}{*}{ViT-B}
    & Uniform     & 0.000 & 0.273 & -0.134 & 0.229 \\
    & Target-only & 0.198 & 0.483 & \underline{0.262}  & 0.201 \\
    & Final-only  & \underline{0.252} & \underline{0.629} & -0.134 & \underline{0.276} \\
    & \textbf{Pairwise} & \textbf{0.259} & \textbf{0.633} & \textbf{0.865} & \textbf{0.358} \\
    
    \hline
    
    \multirow{4}{*}{ViT-L}
    & Uniform     & 0.000 & 0.322 & -0.163 & 0.224 \\
    & Target-only & 0.268 & 0.511 & \underline{0.326}  & 0.193 \\
    & Final-only  & \underline{0.349} & \underline{0.724} & -0.163 & \underline{0.270} \\
    & \textbf{Pairwise} & \textbf{0.355} & \textbf{0.733} & \textbf{0.887} & \textbf{0.344} \\
    
    \hline
    
    \end{tabular}
    \vspace{3pt}
    \caption{Ablation on decision-prior injection under a balanced setting; variants differ only in how the prior is used in propagation.}
    \label{tab:ablation_injection}
\end{table}

As shown in Table~\ref{tab:ablation_injection}, Uniform performs worst overall, indicating that decision prior information is necessary for class-discriminative and propagation-consistent explanations. Target-only improves over Uniform, but remains clearly weaker than propagation-time modulation. Final-only achieves competitive CS and TCC, yet leaves LDA unchanged from Uniform, indicating that post-hoc gating can sharpen the final attribution values without improving the propagation trajectory itself. Pairwise yields the best overall results on both ViT-B and ViT-L, supporting that the main benefit of DAP comes from injecting decision cues directly into token-level propagation rather than applying them only after propagation.

\begin{table}[!t]
    \centering
    \footnotesize
    \renewcommand{\arraystretch}{1.2}
    
    \begin{tabularx}{\textwidth}{c|c|*{5}{>{\centering\arraybackslash}X}|*{5}{>{\centering\arraybackslash}X}}
    \hline
    
    \multirow{2}{*}{\textbf{Model}} & 
    \multirow{2}{*}{\textbf{Method}} & 
    \multicolumn{5}{c|}{\textbf{Balanced Class}} & 
    \multicolumn{5}{c}{\textbf{Non-Balanced Class}} \\
    
    \cline{3-12}
    
    & & 
    \textbf{Del\,{\footnotesize$\downarrow$}} & 
    \textbf{Ins} & 
    \textbf{CS} & 
    \textbf{TCC} & 
    \textbf{AFS} & 
    \textbf{Del\,{\footnotesize$\downarrow$}} &
    \textbf{Ins} & 
    \textbf{CS} & 
    \textbf{TCC} & 
    \textbf{AFS} \\ 
    
    \hline
    
    \multirow{5}{*}{ViT-T}
    & GradCAM & 0.211 & 0.494 & \underline{0.395} & 0.082 & 0.369 & 0.211 & 0.501 & \underline{0.402} & 0.082 & 0.368 \\
    & CDAM & 0.345 & 0.468 & \textbf{0.401} & 0.101 & \textbf{0.828} & 0.343 & 0.476 & \textbf{0.409} & 0.102 & \textbf{0.828} \\
    & AGCAM & \textbf{0.123} & \textbf{0.601} & 0.200 & \textbf{0.172} & 0.216 & \textbf{0.121} & \textbf{0.606} & 0.207 & \textbf{0.168} & 0.215 \\
    & LeGrad & \underline{0.165} & \underline{0.550} & 0.227 & \underline{0.121} & 0.168 & \underline{0.168} & \underline{0.551} & 0.237 & \underline{0.120} & 0.169 \\
    & \textbf{Ours} & 0.200 & 0.510 & 0.386 & 0.117 & \underline{0.525} & 0.200 & 0.519 & 0.392 & 0.107 & \underline{0.523} \\    
    \hline
    \multirow{5}{*}{ViT-S}
    & GradCAM & 0.205 & 0.689 & \textbf{0.316} & 0.339 & 0.320 & 0.205 & 0.693 & \underline{0.320} & 0.325 & 0.319 \\
    & CDAM & 0.478 & 0.610 & \underline{0.314} & 0.206 & \textbf{0.706} & 0.486 & 0.612 & \textbf{0.325} & 0.227 & \textbf{0.717} \\
    & AGCAM & \textbf{0.195} & \textbf{0.703} & 0.158 & \textbf{0.392} & 0.214 & \textbf{0.193} & \textbf{0.708} & 0.164 & \textbf{0.383} & 0.215 \\    
    & LeGrad & \underline{0.199} & 0.693 & 0.179 & 0.371 & 0.170 & \underline{0.199} & 0.697 & 0.190 & \underline{0.379} & 0.171 \\
    & \textbf{Ours} & 0.209 & \underline{0.696} & 0.306 & \underline{0.380} & \underline{0.445} & 0.206 & \underline{0.701} & 0.309 & 0.372 & \underline{0.441} \\
    \hline
    \multirow{5}{*}{ViT-B}
    & GradCAM & 0.248 & 0.748 & \underline{0.262} & 0.570 & 0.252 & 0.259 & 0.761 & \underline{0.257} & 0.616 & 0.250 \\
    & CDAM & 0.484 & 0.745 & \textbf{0.303} & 0.549 & \textbf{0.675} & 0.494 & 0.755 & \textbf{0.302} & 0.579 & \textbf{0.675} \\
    & AGCAM & \textbf{0.235} & \textbf{0.759} & 0.110 & \textbf{0.673} & 0.217 & \textbf{0.240} & \textbf{0.774} & 0.107 & 0.696 & 0.218 \\
    & LeGrad & \underline{0.242} & 0.754 & 0.146 & \underline{0.648} & 0.172 & \underline{0.253} & 0.769 & 0.145 & \textbf{0.679} & 0.172 \\
    & \textbf{Ours} & 0.249 & \underline{0.755} & 0.259 & 0.633 & \underline{0.358} & 0.259 & \underline{0.770} & 0.252 & \underline{0.673} & \underline{0.353} \\
    \hline
    \multirow{5}{*}{ViT-L}
    & GradCAM & 0.383 & 0.795 & \underline{0.355} & 0.717 & 0.258 & 0.387 & 0.793 & \textbf{0.369} & 0.754 & 0.256 \\
    & CDAM & 0.728 & 0.659 & 0.214 & 0.327 & \textbf{0.855} & 0.731 & 0.665 & 0.208 & 0.329 & \textbf{0.856} \\
    & AGCAM & \underline{0.379} & 0.794 & 0.126 & 0.703 & 0.192 & \underline{0.383} & 0.786 & 0.122 & 0.731 & 0.192 \\
    & LeGrad & \textbf{0.343} & \textbf{0.805} & 0.132 & \textbf{0.773} & 0.180 & \textbf{0.345} & \textbf{0.800} & 0.132 & \textbf{0.802} & 0.180 \\
    & \textbf{Ours} & 0.398 & \underline{0.798} & \textbf{0.355} & \underline{0.733} & \underline{0.344} & 0.398 & \underline{0.795} & \underline{0.367} & \underline{0.756} & \underline{0.342} \\
    \hline
    
    \end{tabularx}
    \vspace{3pt}
    \caption{Comparison of the proposed method with gradient-based XAI approaches across multiple ViT models under both balanced and non-balanced class settings}
    \label{tab:metric_grad}
\end{table}

\subsubsection{Comparison with Gradient-based Methods}

Relative to gradient-based methods (Table ~\ref{tab:metric_grad}), the proposed method remains competitive despite not being a purely gradient-driven explanation method. Although it did not consistently outperform the leading baselines, especially on Del, Ins, and TCC, it still produced favorable results while preserving transformer attention propagation. This difference is expected, as gradient-based methods focus more directly on class-discriminative localization, whereas the proposed approach also preserves the layer-wise attention structure of the transformer. Nevertheless, it maintained relatively high CS scores on larger backbones, reaching 0.259 on ViT-B and 0.355 on ViT-L. This is meaningful because it shows that gradient-based decision cues can be incorporated into an attention-aware explanation framework while retaining competitive discriminative performance. Overall, these findings suggest that the proposed method captures important gradient-based characteristics without discarding the structural advantages of attention-based explanation, thereby providing a reasonable balance between class sensitivity and propagation-based faithfulness.

\begin{figure}
  \centering
  \includegraphics[width=\textwidth]{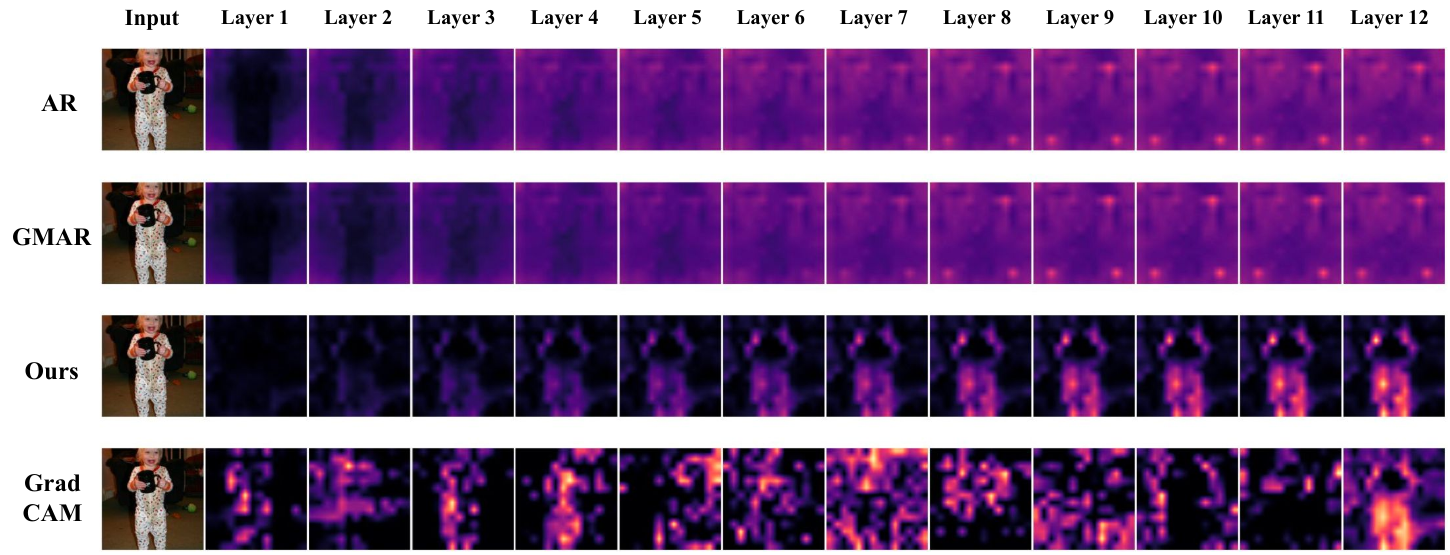}  
  \caption{Layer-wise Attention Map Comparison Across Methods}
  \label{fig:layer}
\end{figure}

\subsection{Qualitative Analysis}

Figure~\ref{fig:layer} shows how class-relevant evidence evolves across layers for different explanation methods. DAP exhibits a more consistent progression toward the target class, progressively strengthening responses on semantically relevant regions while preserving a coherent object-level structure across depth. In contrast, Attention Rollout and GMAR produce broader and more diffuse responses, often spreading over surrounding context rather than concentrating on the most decision-relevant evidence. Grad-CAM eventually highlights class-relevant regions, but its layer-wise activations are less stable and appear more fragmented across depth. Overall, these visualizations suggest that DAP yields explanations that are both more class-consistent and more sequentially coherent across transformer layers.

\subsection{Explanation Property Analysis}

\begin{figure}[h]
  \centering
  \includegraphics[width=\textwidth]{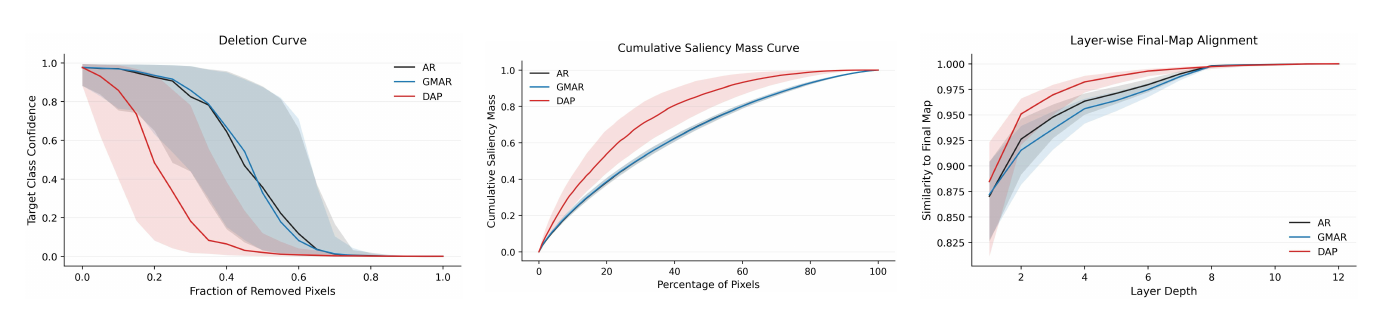}  
  \caption{Evaluation of Explanation Quality via Deletion, Mass, and Alignment Curves}
  \label{fig:graph}
\end{figure}

Figure~\ref{fig:graph} analyzes explanation quality from three complementary perspectives. In the deletion curve, better explanations cause the target confidence to drop more rapidly as salient regions are removed, and DAP shows the steepest decline among the compared methods, indicating that its highlighted regions are more directly tied to the final prediction. In the cumulative saliency mass curve, stronger performance is reflected by faster early accumulation, and DAP concentrates relevance more quickly than the baselines, suggesting a more compact attribution pattern. In the layer-wise final-map alignment curve, better explanations maintain higher similarity to the final map even at earlier layers, and DAP remains more strongly aligned throughout depth, indicating more consistent propagation toward the final decision. Taken together, these results show that DAP improves decision sensitivity, compactness, and propagation consistency relative to the baselines.

\section{Conclusion}
\label{sec:con}

DAP was proposed to improve ViT explainability by integrating gradient-derived decision cues into attention propagation, with the goal of preserving transformer-specific attention flow while enhancing class-discriminative relevance. The experimental results showed that this strategy was particularly effective on larger ViT backbones, where it improved class sensitivity, token contribution consistency, propagation-level alignment, and layer-wise attribution stability. These findings suggest that directly incorporating decision-aware token importance into propagation provides a practical way to better balance faithfulness to internal attention structure and discrimination for the final prediction. Although the current study was limited to ImageNet-pretrained ViTs and did not examine other datasets or transformer-based vision architectures, the proposed framework offers a promising direction for more reliable and interpretable transformer explanations in real-world applications.

\bibliographystyle{unsrtnat}
\bibliography{ref}

\newpage
\appendix

\begin{center}
{\Large\bfseries Appendix}
\end{center}
\vspace{1em}

\renewcommand{\thefigure}{A.\arabic{figure}}
\renewcommand{\thetable}{A.\arabic{table}}
\setcounter{figure}{0}
\setcounter{table}{0}

\section{Detailed Metric Definitions}
\label{app:metrics}

This appendix provides implementation details and additional interpretation for the evaluation metrics used in the main paper. While the main text focuses on concise definitions, here we clarify how each metric is computed in practice for token-based Vision Transformer explanations and what aspect of explanation quality it is intended to capture.

\paragraph{Class Sensitivity (CS).}
CS measures whether an explanation changes when the queried class changes. For each input image, we generate one explanation map for the predicted class and another for an alternative class. Let
\[
e^{\mathrm{pred}}, e^{\mathrm{alt}} \in \mathbb{R}^{P}
\]
denote the corresponding patch-level explanation vectors after flattening the token grid. We compute the Spearman rank correlation between the two vectors and convert it into a separation score:
\[
\mathrm{CS} = 1 - \frac{\rho_s(e^{\mathrm{pred}}, e^{\mathrm{alt}})}{2}.
\]
A high CS indicates that the patch ranking changes substantially across target classes, meaning that the explanation is more class-discriminative. Conversely, a low CS indicates that similar patch importance patterns are produced even when the queried class changes.

\paragraph{Token Contribution Consistency (TCC).}
TCC evaluates whether the patches highlighted by an explanation are sufficient to preserve the original prediction. Given a patch-level explanation
\[
e = [e_1, \dots, e_P],
\]
we select the top-$k$ patches according to their relevance values, where $k=10\%$ in our experiments:
\[
\mathcal{K} = \mathrm{TopK}(e).
\]
Because explanations are defined on the token grid while perturbation is applied in image space, the selected patch indices are mapped back to their corresponding spatial regions in the input image. This produces a binary mask
\[
m(u,v) \in \{0,1\},
\]
where $(u,v)$ denotes the pixel location and $m(u,v)=1$ if the pixel belongs to one of the selected patches. The masked input is then defined as
\[
x_{\mathrm{top}k}(u,v)=x(u,v)\,m(u,v).
\]
TCC compares the target-class confidence before and after masking:
\[
\mathrm{TCC}=\frac{p(y_t \mid x_{\mathrm{top}k})}{p(y_t \mid x)}.
\]
A high TCC indicates that the selected patches retain a large fraction of the evidence used for the original decision. This metric therefore reflects the \emph{sufficiency} of the highlighted regions rather than their compactness or class selectivity.

\paragraph{Patch Selection and Patch-to-Image Mapping.}
Several metrics in this work operate on patch-level explanation maps but require either perturbation or aggregation in image space. To ensure consistency, top-$k$ selection is always performed first on the patch grid. The selected patch indices are then expanded to their corresponding image regions according to the ViT patch partition. In TCC, this produces the binary mask used for image-space perturbation. In this way, the evaluation remains faithful to the token-level explanation while still allowing perturbation to be applied to the original image.

\paragraph{Attention Flow Sparsity (AFS).}
AFS measures how strongly relevance is concentrated on a small subset of patches. Given a nonnegative patch-level explanation map
\[
e \in \mathbb{R}^{P},
\]
we compute the fraction of total relevance mass contained in the top-$k$ patches:
\[
\mathrm{AFS}_{\mathrm{top}k}
=
\frac{\sum_{i \in \mathrm{TopK}(e)} e_i}{\sum_j e_j}.
\]
A high AFS indicates that relevance is concentrated on a relatively small set of tokens, corresponding to a more compact explanation. A low AFS indicates that relevance is spread more broadly over the token grid. Accordingly, AFS should be interpreted as a measure of concentration rather than as a direct measure of correctness.

\paragraph{Layer-wise Decision Alignment (LDA).}
LDA evaluates how consistently intermediate relevance maps develop toward the final explanation. Let
\[
m^{(l)} \in \mathbb{R}^{P}
\]
denote the patch-level relevance map obtained after propagation up to layer $l$, and let
\[
m^{(\mathrm{ref})}
\]
denote the final-layer attribution map produced by the same method. We compute
\[
\mathrm{LDA}_l = \rho_s\!\left(m^{(l)}, m^{(\mathrm{ref})}\right),
\]
where $\rho_s$ is the Spearman rank correlation. Under this formulation, LDA does not evaluate whether an intermediate map is absolutely correct with respect to an external annotation. Instead, it measures whether the relevance pattern at layer $l$ is already evolving in a direction consistent with the final decision map. Higher LDA therefore indicates stronger internal consistency of layer-wise propagation.

\paragraph{Choice of Top-$k$.}
For all top-$k$-based evaluations, we fix $k=10\%$ of the patch tokens in the main experiments. This ratio is used consistently in both TCC and AFS so that sufficiency and concentration are evaluated under the same token budget.

\section{Additional Comparison of Gradient-Integrated Attention Methods}
\label{app:grad_integrated}

To provide a more focused view of methods that combine gradient cues with attention-based explanations, Table~\ref{tab:metric_gmar_gradcam} compares DAP with GMAR and Grad-CAM on ViT-B and ViT-L. This comparison is supplementary to the main quantitative results in the paper, where the full attention-based and gradient-based comparisons are presented separately.

Compared with GMAR, DAP shows clear gains on both ViT-B and ViT-L, especially in CS, TCC, and AFS. This trend is consistent with the design difference between the two methods: GMAR uses gradients to reweight attention heads before rollout, whereas DAP injects decision cues directly into token-level propagation. Compared with Grad-CAM, DAP remains competitive in Del and CS while showing stronger propagation-aware behavior and favorable TCC/AFS values on both backbones. These results further support the view that directly modulating token-level relevance propagation provides a more effective integration of gradient-derived decision cues and transformer attention structure than head-level weighting alone.

\begin{table}[ht]
    \centering
    \renewcommand{\arraystretch}{1.2}
    
    \begin{tabular}{llccccc}
    \hline
    \textbf{Method} & \textbf{Model} & \textbf{Del $\downarrow$} & \textbf{Ins} & \textbf{CS} & \textbf{TCC} & \textbf{AFS} \\
    \hline
    
    GMAR & ViT-B & 0.396 & 0.653 & 0.000 & 0.273 & 0.231  \\
    Grad-CAM & ViT-B & \textbf{0.248} & 0.748 & \textbf{0.262} & 0.570 & 0.252  \\
    \textbf{Ours} & ViT-B & 0.249 & \textbf{0.755} & 0.259 & \textbf{0.633} & \textbf{0.358}  \\
    
    \hline\hline
    
    GMAR & ViT-L & 0.565 & 0.686 & 0.000 & 0.292 & 0.225  \\
    Grad-CAM & ViT-L & \textbf{0.383} & 0.795 & 0.355 & 0.717 & 0.258 \\
    \textbf{Ours} & ViT-L & 0.398 & \textbf{0.798} & \textbf{0.355} & \textbf{0.733} & \textbf{0.344}  \\
    
    \hline
    \end{tabular}
    \vspace{6pt}
    \caption{Supplementary comparison of DAP with GMAR and Grad-CAM on ViT-B and ViT-L. This table provides a focused view of gradient-integrated attention explanations and complements the broader comparisons.}
    \label{tab:metric_gmar_gradcam}
\end{table}

\begin{figure}[p]
  \centering
  \begin{subfigure}{\textwidth}
    \centering
    \includegraphics[width=0.8\linewidth]{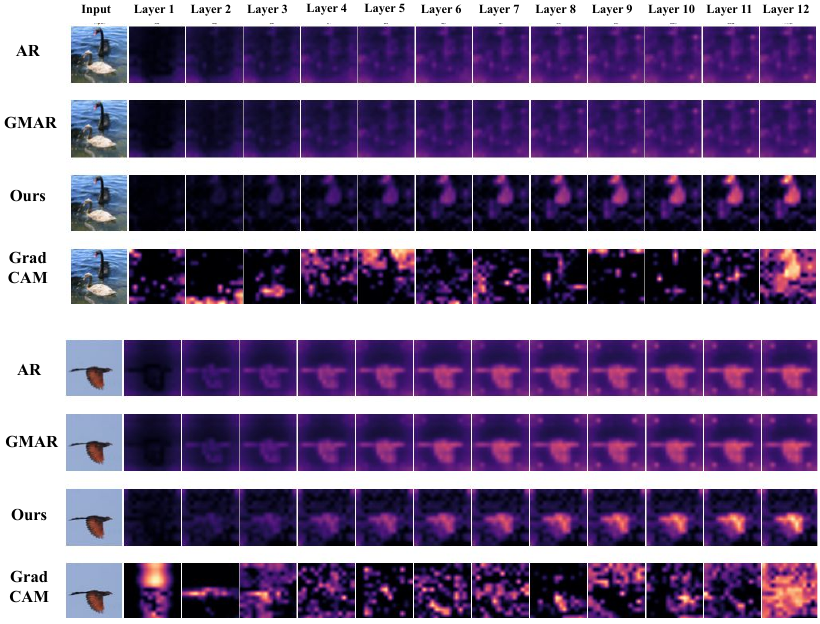}
    \caption{Successful Case}
    \label{fig:good}
  \end{subfigure}
  
  \vspace{1cm} 

  \begin{subfigure}{\textwidth}
    \centering
    \includegraphics[width=0.8\linewidth]{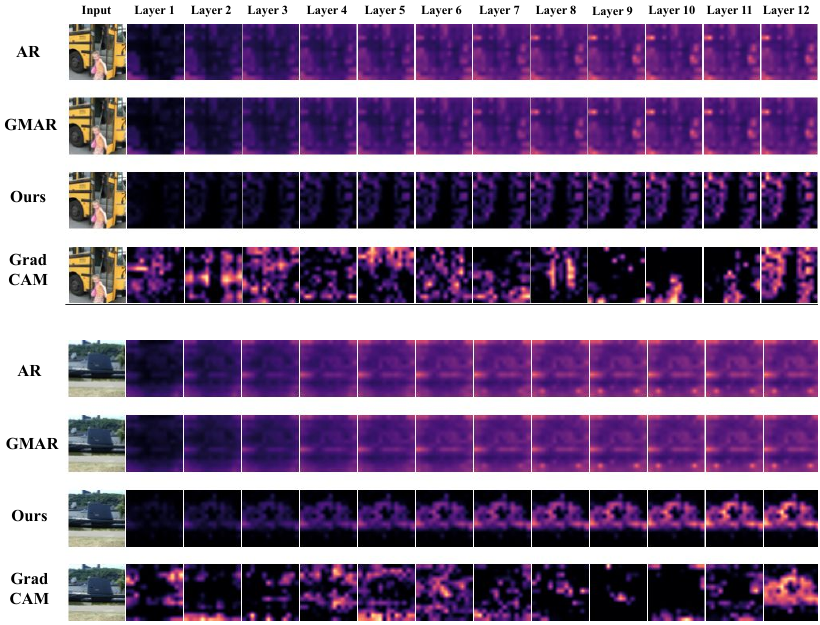}
    \caption{Challenging Case}
    \label{fig:bad}
  \end{subfigure}
  
  \caption{Additional Layer-wise Attention Maps for Successful and Challenging Cases}
  \label{fig:comparison}
\end{figure}

\section{Additional Qualitative Results}

To further examine the behavior of the proposed method across different visual conditions, we provide additional layer-wise attention visualization examples. Specifically, we include a representative successful case and a more challenging case, shown in the Figure~\ref{fig:comparison}. The successful case demonstrates that the proposed method maintains a relatively stable and class-relevant attention flow across layers, gradually strengthening responses on the target object while suppressing less relevant regions. In contrast, the failure case shows that this behavior can become weaker when the target evidence is less distinctive or when the scene contains more ambiguous visual patterns. These examples complement the main-paper qualitative results by illustrating both the strengths and the limitations of the method under different levels of difficulty.

In the successful examples (Figure~\ref{fig:good}), Attention Rollout and GMAR tend to produce broadly diffused responses over the scene, which makes it harder to identify when the explanation becomes clearly aligned with the final prediction. Grad-CAM provides more class-discriminative responses, but its activations appear less consistent across layers and often fluctuate between scattered regions. By contrast, the proposed method shows a more coherent progression, where class-relevant regions emerge earlier and remain more consistently emphasized as the network depth increases. This suggests that the method better preserves layer-wise propagation behavior while guiding the attention flow toward decision-relevant evidence.

In the challenging examples (Figure~\ref{fig:bad}), the proposed method still exhibits a more structured propagation trend than the baselines, but the final focus becomes less clean and less selective than in the successful case. This indicates that although the decision-aware propagation mechanism improves consistency and discrimination in many cases, it does not fully resolve ambiguity when the object boundary is unclear, when multiple visually plausible regions coexist, or when the model itself lacks sufficiently strong class evidence. We include these cases to provide a more balanced qualitative assessment and to show that the method improves explanation behavior in a consistent direction, while still inheriting limitations from the underlying backbone prediction.

\subsection{Equivalence of Source-Only and Pairwise Injection Under Row Normalization}

For completeness, we briefly clarify why the source-only variant is not reported separately in the main ablation. Let $\hat{\mathbf{A}}^{(l)} \in \mathbb{R}^{N \times N}$ denote the residual-aware attention matrix at layer $l$, and let $\mathbf{d} \in \mathbb{R}^{N}$ be the token-level decision prior. In the pairwise form used in DAP, the transition score from token $j$ to token $i$ is modulated as
\begin{equation}
\tilde{A}^{(l)}_{ij} = \hat{A}^{(l)}_{ij} \, d_i d_j .
\end{equation}
After row normalization, the actual propagation matrix becomes
\begin{equation}
P^{(l)}_{ij}
=
\frac{\hat{A}^{(l)}_{ij} \, d_i d_j}
{\sum_{k} \hat{A}^{(l)}_{ik} \, d_i d_k }.
\end{equation}
Since $d_i$ is shared by all entries in row $i$, it cancels out:
\begin{equation}
P^{(l)}_{ij}
=
\frac{\hat{A}^{(l)}_{ij} \, d_j}
{\sum_{k} \hat{A}^{(l)}_{ik} \, d_k }.
\end{equation}
Therefore, the normalized pairwise form is identical to applying the prior only to the source token:
\begin{equation}
\tilde{A}^{(l)}_{ij} = \hat{A}^{(l)}_{ij} \, d_j .
\end{equation}
This shows that, under row normalization, the target-side factor does not affect the final transition probability, and the pairwise and source-only variants become mathematically equivalent. For this reason, only one of these forms is necessary in the ablation.

\section{Link of Our Codes}
Code is available at https://anonymous.4open.science/r/dap-395D

\section{Limitations}
This method was evaluated only on ImageNet-pretrained ViTs, which limits the validation of its generalizability across datasets and model architectures. In particular, its effectiveness was not examined on datasets with different characteristics or on other transformer-based vision models beyond ViT. Future work will extend the evaluation to more diverse datasets and a broader range of transformer architectures to further verify the robustness and general applicability of the proposed method.

\section{Societal impact}
Improving the explainability of ViTs can enhance the reliability of their use in real-world applications. By providing more transparent and interpretable evidence for model predictions, the proposed approach can help users better understand and trust the decision-making process of ViT-based systems. This is particularly important in high-stakes domains such as healthcare, where the justification for a model’s prediction is often as important as the prediction itself. In such settings, offering explanation results that users can reasonably interpret and accept may support the more effective and responsible deployment of ViT models in practice.


\end{document}